# Predicting IMDb Rating of TV Series with Deep Learning: The Case of Arrow

Anna Luiza Gomes
Informatics Department
PUC-Rio
Rio de Janeiro, Brazil
annalgs@id.uff.br

Getúlio Vianna Jr.
Informatics Department
PUC-Rio
Rio de Janeiro, Brazil
getulio.a.vianna@gmail.com

Tatiana Escovedo
Informatics Department
PUC-Rio
Rio de Janeiro, Brazil
tatiana@inf.puc-rio.br

Marcos Kalinowski
Informatics Department
PUC-Rio
Rio de Janeiro, Brazil
kalinowski@inf.puc-rio.br

## ABSTRACT

**Context:** The number of TV series offered nowadays is very high. Due to its large amount, many series are canceled due to a lack of originality that generates a low audience.

**Problem:** Having a decision support system that can show why some shows are a huge success or not would facilitate the choices of renewing or starting a show.

**Solution:** We studied the case of the series Arrow broadcasted by CW Network and used descriptive and predictive modeling techniques to predict the IMDb rating. We assumed that the theme of the episode would affect its evaluation by users, so the dataset is composed only by the director of the episode, the number of reviews that episode got, the percentual of each theme extracted by the Latent Dirichlet Allocation (LDA) model of an episode, the number of viewers from Wikipedia and the rating from IMDb. The LDA model is a generative probabilistic model of a collection of documents made up of words.

**IS Theory:** This study was developed under the aegis of Computational Learning Theory, which aims to understand the fundamental principles of learning and contribute to designing better-automated learning methods applied to the entertainment business.

**Method:** In this prescriptive research, the case study method was used, and its results were analyzed using a quantitative approach.

**Summary of Results:** With the features of each episode, the model that performed the best to predict the rating was Catboost due to a similar mean squared error of the KNN model but a better standard deviation during the test phase. It was possible to predict IMDb ratings with an acceptable root mean squared error of 0.55.

**Contributions and Impact in the IS area:** The results show that deep learning techniques can be applied to support decisions in the entertainment field, allowing facilitating the decisions of renewing or starting a show. The rationale for building the model is detailed throughout the paper and can be replicated for other contexts.

## CCS CONCEPTS

• Information systems • Information systems applications • Enterprise information systems • Enterprise applications

## KEYWORDS

Machine Learning, Deep Learning, LDA, prediction, tv-series, IMDb.

## 1 Introduction

The number of people who consume TV series has grown a lot over the years, as well as the number of series offered. Netflix, for example, has more than 183 million global subscribers, measured in 2020 [3]. Many television productions that are canceled cause loss of investments or delays in other possible projects that would yield greater financial return, and in the majority of the cases, it is due to the low audience. Understanding what the public likes are the primary goal of its line of sight work.

There are several studies on the success forecast of series and films. Some analyze the reaction of the public on social media and others the features (script, director, actors, actresses, genre, etc.) that influence its popularity. There are platforms where it is possible to discover the ratings of the public or critics about a series, such as IMDb (Internet Movie Database), Rotten Tomatoes, Metacritic, among others, which help to get a sense of public perception.

According to research conducted by FX, there was a record of 532 scripted TV programs in 2019 [4]. That said, it is important to not overwhelm the TV schedule with programs that don't differ much, leading to a waste of time for viewers, artists, producers, or anyone involved.

Machine Learning is a subfield of Artificial Intelligence closely related to applied mathematics and statistics and has many applications in everyday life [10]. For example, predicting the IMDb rating of a series is a regression task encompassed by Machine Learning techniques. We can define a regression problem as given a set of $n$ patterns, each one consisting of information of explanatory variables (independent) and a variable continuous or discrete (dependent) response. The goal is to build a regression model that, given a new pattern, estimates the most expected value for the response variable [18].

Linear regression is an example of an analytical technique used to model the relationship between several input variables and a continuous outcome variable [11]. The model (an abstraction of reality) created provides insight into the underlying processes of a phenomenon [10], which in this case is the rating of a series.

With Machine Learning techniques, it is also possible to identify the episode's main topics through the script and analyze other possible relevant characteristics that impact its success with the public. Thus, enabling data-driven decisions of projects that might bring a better return of investments being more confident whenever sending projects to production.

Therefore, the main objective of this work is to develop a model that can efficiently predict the IMDb classification of the Arrow series. Arrow is an American TV show broadcasted by CW Network, released in 2012. It aired its last episode in 2020; this series has been chosen for this study by convenience due to the author's personal taste and knowledge regarding the content of its episodes. It's based on a comic book, and the genre is a mix of action, adventure, crime, drama, mystery, and Sci-Fi. IMDb is the chosen platform for its popularity and familiarity.

Our results indicate that it was possible to build a model that can predict IMDb ratings with an acceptable root mean squared error of 0.55 (i.e., the predicted ratings differ from the real ratings in about half a point in a scale from 0 to 10). The model that performed the best to predict the rating was Catboost due to a similar mean squared error of the KNN model but a better standard deviation during the test phase.

The following is an outline of the paper. First, we describe the related work of TV series rating prediction and a brief overview of Latent Dirichlet Allocation. The next section describes our experimental methodology, including our methods for extracting, filtering, and cleaning our dataset before applying topic modeling. Also, we present our topic modeling results for each episode of Arrow and analysis. Finally, we conclude with a discussion and next steps.

## 2 Related Work

We can find several works in literature to predict movie and series success rates using its features. The research has been done on predicting the rating of movies using its information like the studio, director, screenplay, actor, actress, genre, and the country as input features and, as the preprocessing step was completed, several classifiers (Bagging, Random Forest, J48, IBK, and Naive Bayes) was used in a repeated style using Weka [1].

Other than films, TV shows, music shows, etc., can be predicted by their model, using its features: title, studio, director, screenplay, actor, actress, genre, country, year.

In [6], the ratings assigned by viewers and the number of reviewers (voters) are adopted as a measure of performance of TV series. The features for prediction are season, episode number within the season; text length; the complexity of the text; language sentiment; use of function words, the concentration of dialogues, and gender of voters.

The correlation between the scripts (the text characteristics) and the success of the series was analyzed. It was shown that the more articulated the episode was, the more popular it got, and higher was the ratings, as suggested by the significant and positive correlations with several words and complexity. They have identified the most relevant predictors of popularity and appreciation through correlation and regression analysis.

In [7], five TV series were analyzed to determine what makes a sitcom successful. This work used descriptive statistics, visualizations, hypothesis testing, and predictive analytics to understand the differences and impacts those distinct factors have on episode rating. The factors were: title, director, writer, original air date, rating, and how many lines each character spoke. The character presence, director, and writer are shown to be accurate predictors of a show's IMBd ratings. It points out that there is no one-size-fits-all model for predicting the rating, possibly due to the immense number of factors that go into producing a television show.

An important text mining phase is determining the main topics for each episode under study. Thus, the algorithm chosen in this work is the most common methodology for performing this kind of job, known as LDA (Latent Dirichlet Allocation). The first time presented in [8] wherefrom unstructured data, mainly text, it is possible to label a topic for each word or the basic unity of each collection of documents, then a percentage topic distribution of each document is achieved. They also assist in discovering some semantics for each topic, and this methodology is intensively used for processing a large number of documents.

There are several applications of this methodology, and [9] had presented those with huge research through the literature. There are applications in medical sciences, software engineering, geography, political science, social media, image processing, crime science, computer vision, etc. For example, LDA can be used to facilitate exploring the dynamic of the interaction of the communities in social media. It can also be used in the case of software engineering to identify the main topics on the source code repository. Another application is finding the most common topics of political speeches over time.

Topic modeling is an unsupervised machine learning technique similar to clusterization of non-text data that aims to find natural groups of items even if there is no defined goal. In this context, LDA is a method for fitting a topic model and one of the most common algorithms for topic modeling, describing a document as a mixture of topics where each topic is a mixture of words. It is a mathematical method that estimates the mixture of words associated with each topic while also estimating the mixture of topics of each document [2].

A document is a mixture of topics, and a topic is a discrete probability distribution that defines the likelihood of each word to be within a topic. A document can be defined by a "bag of the words" without structure, only having a topic assigned and word statistics [5].

The main assumption of this paper is that there is a relation between the episodes' main subjects addressed with IMDB's grade, and due to that, the Topic modeling approaches have been chosen as the candidate to retrieve the subjects and semantics from each episode.

## 3 Proposed Method

Our research consists of four main steps: data collection through web scraping, data cleaning, implementation of text mining approaches (Word Cloud and Topic Modeling), and regression to predict the IMDb rating.

### 3.1 Problem Description

In this work, we aim to predict the episode's rating on IMDb based on its scripts and relevant information, such as the director of the episode, the number of viewers, and the number of reviewers on IMDb. The model for the prediction is built using machine learning techniques. It is trained and tested by the dataset extracted from Wikipedia, IMDb series rating website, and Subslikescript. The script was scraped from the Subslikescript section of Arrow, available on < https://subslikescript.com/series/Arrow-2193021>. From Wikipedia, we obtained information about the director and the number of viewers of every episode.

The rating and the number of reviewers were obtained manually from the IMDb site because the site has restrictions on using web crawlers.

### 3.2 Data Description

Our dataset is formed by: the director of the episode, the number of reviews on IMDb, the score on each topic (through the LDA model), the predominant topic, the number of viewers on Wikipedia, and the rating on IMDb. The complete dataset has 165 episodes (rows) and seven columns in total.

All the information are numbers except the episode's director, so it was necessary to transform this column to a number to facilitate the modeling phase. As this column had 51 unique names, we chose an encoder with just one output dimension, and as the names are independent of any other column, the Ordinal Encoder. This process assigns numbers for each unique value in the column.

LDA topic modeling estimates two sets of distributions: 1) representing each document as a multinomial distribution over T topics and 2) representing each topic with a multinomial distribution over W words [15]. Being an unsupervised model, once we provide the algorithm with the number of topics, it analyses the documents and applies a topic for each to obtain a good composition of the topic-keywords distribution. A topic is a collection of dominant words representing a subject or content.

First, a Word Cloud was generated to assess the more proeminent in the dataset to remove the less important ones, as seen in Figure 1. Alongside the stopwords in English were added some other words that repeated themselves and did not help to define a topic, such as: "from", "arrow", "oh," etc.

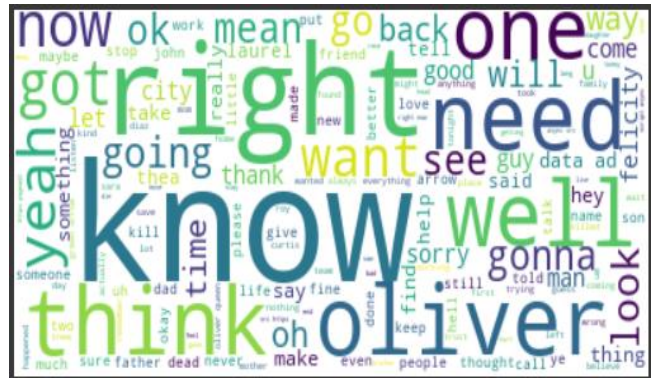

**Figure 1: IMDB rating histogram**

The LDA output for each row is a sum of 1 for all three clusters because the episode is proportionally labeled for each cluster for its words. Therefore, the values of the three clusters range from om 0 to 1. Wikipedia viewers are shown in millions ranging from 0.62 to 4.14 with a standard deviation of 0.86.

The IMDB grade ranges from 5.5 to 9.7 and standard deviation of 0.64; important to note that more than 50% of grades are between 8.0 and 9.0, as shown in Figure 2. The outcome variable is the IMDB grades, and the independent variables are the remaing ones.

After analyzing the IMDb grades histogram, the correlation between the independent variables and the dependent variables was performed using Pearson's correlation method. The correlation value is a number between -1 and 1, that values with no correlation are near zero and values near +1 and -1 are highly correlated, positive values correspond that both compared variables increase or decrease commonly, and negative values correspond that the change in the opposite direction.

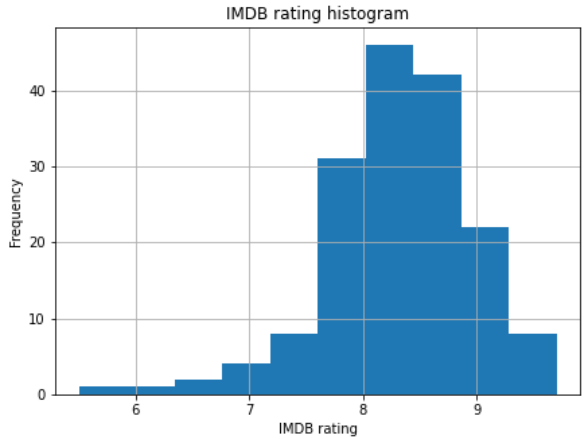

**Figure 2: IMDB rating histogram**

From Pearson's correlation (shown in Figure 3), it is possible to notice that all independent variables (X) have a low correlation with the dependent variable (Y). They range from an absolute value between 0.012 and 0.38.

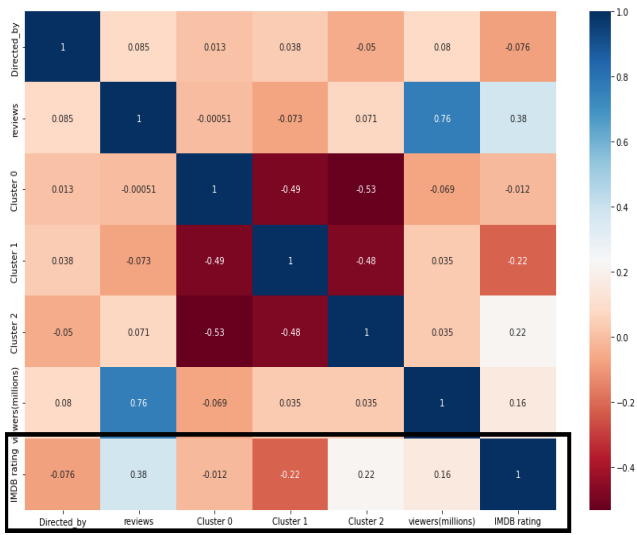

**Figure 3: Pearson's correlation matrix**

As previously mentioned, the number of reviews is much bigger than other variables, so all independent variables were normalized during training and tests, and the predictor (Y) was kept without transformation. MinMaxScaler has been chosen as a transformation algorithm. For each column, it takes the lowest and highest values to be 0 and 1, respectively, and the other values to fit in between those values.

### 3.3 Proposed Methodology and Results

One approach to solving the problem of identifying the main topic of an episode is topic modeling. For the rating prediction, a linear regression model was used.

The percentage of adherence to each topic generated by the LDA model in each episode was retrieved to understand its composition. Before feeding the algorithm with data (scripts of all episodes), it was necessary to remove the stopwords and the unnecessary characters. For example, there was a series of introduction passphrases in most episodes, which was also removed from the dataset.

The lemmatization technique - that in linguistics is the process of grouping together the inflected forms of a word to be analyzed as a single item, identified by the word's lemma - was also applied. For example, it transforms the words working, works, worked into the normalized form work standing for the infinitive: work.

The algorithm was configured to generate three topics because it was the most explainable way to define the topic and the better way to separate into different clusters of keywords, shown in Figure 4. The size of topics represents their distribution across the dataset (165 episodes), and the more distant they appear, the more different they are from each other, just like clusterization.

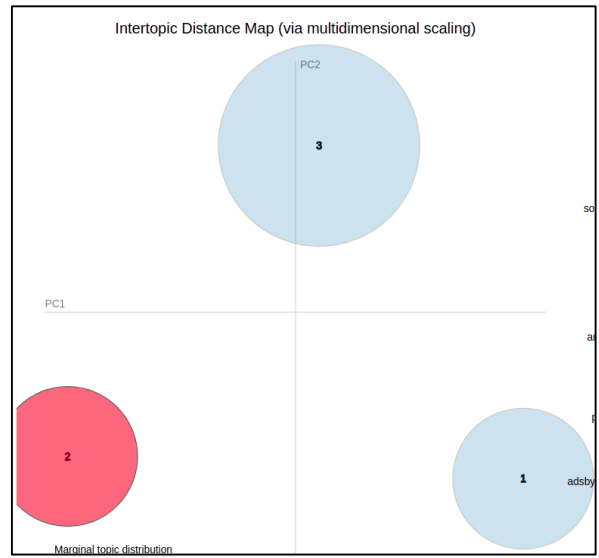

**Figure 4: The topics generated by the LDA model and its distribution**

The topics were defined by the authors through knowledge of the series and its characters, in the list below:

1. Topic 0 – It's an episode more **generalistic, focused on action,** because of the inclusion of other characters that are the main character's friends, family, and romantic interests. In addition to being protagonists in action scenes. It also includes the word kill as one of the most frequent in this cluster.

2. Topic 1 - It's an episode focused on **family plots** more than the others because it lacks the word Kill or others that resemble action scenes. Also, the sister (Thea) and the father come before the first romantic interest of the main character, Laurel.

3. Topic 2 – It's an episode more **romantic** because the two biggest romantic pairs of the main character (Oliver), Felicity and Laurel, are on the keywords list and second in order of frequency.

The main keywords are shown in Figure 5.

```
[(0,
 '0.009*"oliver" + 0.004*"felicity" + 0.003*"help" + 0.003*"laurel" + '
 '0.003*"please" + 0.003*"thea" + 0.002*"john" + 0.002*"father" + '
 '0.002*"kill" + 0.002*"give"'),
 (1,
 '0.007*"oliver" + 0.004*"felicity" + 0.003*"help" + 0.003*"thea" + '
 '0.002*"father" + 0.002*"laurel" + 0.002*"stop" + 0.002*"friend" + '
 '0.002*"someone" + 0.002*"give"'),
 (2,
 '0.009*"oliver" + 0.004*"laurel" + 0.003*"felicity" + 0.003*"help" + '
 '0.003*"thea" + 0.003*"father" + 0.002*"kill" + 0.002*"friend" + '
 '0.002*"john" + 0.002*"last"')]
```

**Figure 5 – Topics generated by the LDA model and its keywords**

After getting the percentual of each topic for each episode, we set a web crawler to get the number of viewers and the episode's director listed on Wikipedia to merge in the dataset. The last additions to the dataset were the number of reviewers column and the IMDb rating.

In order to avoid bias during training phases, all models have been trained using k fold cross-validation [17], which means that the dataset has been split into almost equally or equally K subsets. Training is performed with K-1 subsets, and the test is performed on the subset removed at the beginning. The mean of standard deviation and root mean squared error is calculated for comparison.

It was decided to begin with models of low complexity. Then a higher complexity could be chosen in sequence. We started with linear regression [12], which mainly has the objective to fit a linear model, get the coefficients, and then calculate the sum of squares between observed and predicted values. In this case, it is a multiple linear regression because it models the relation of more than one independent variable with the predictor.

In sequence, we performed tests with KNN [13], which means K-nearest neighbors. This algorithm assumes that a group of K near neighbors share similar values to its group. The process relies on distance measurement between the points being studied. It can be used for classification problems that the difference is that the similar attributes of near neighbors correspond to a specific class.

On the other hand, regression problems consist of evaluating the average value of the nearest neighbors.

Catboost [14] was the third model used and is a boosting technic of ensemble models. In general, it works on building in series weak symmetric decision trees with permutation samples from the whole dataset. The residuals (errors) of previous decision trees are used as the target for the subsequent decision tree. In the end, it makes an average of the whole decision trees prediction created.

The training procedure was done with 80% of the dataset (132 rows), and the test dataset consisted of the remaining 20% (33 rows). We tested during the training phase the usage of the raw and the normalized dataset. Still, the normalized was always slightly better than raw, so all the results shown are considering using normalization to independent variables (X) and raw values for predictor (Y). The metric used for all tests and training evaluations was the root mean squared error (RMSE).

For all training and test phase models, the dataset used was normalized for independent variables and with no transformation for the dependent variable. The cross-validation k-fold used was 10 for all models during the training phase.

Linear regression model training was performed followed by KNN regressor with a Grid search method that sought the best parameters for the model around the following hyperparameters range: k neighbors ranging from 1 to 16 and algorithm' ball tree'. The best parameter was k = 13—no change on algorithm due to neglectable changing.

The last model trained was Catboost. The same metric (RMSE) was used as a loss function for the model assessment. A grid search was performed around the following parameters: learning rate of 0.03, 0.1, and 0.2; depth of 4, 6, 10, 20, and 30; l2_leaf_reg of 1, 3, 5, 7, 9, 12, 15. The best parameters achieved were depth of 6, l2_leaf_reg of 7, and learning_rate of 0.1. Table 1 shows the results for the training phase.

**Table 1: Training models results**

| Training phase – Best Models | RMSE | Standard Deviation |
|---|---|---|
| Linear Regression | 0.5602 | 0.2288 |
| KNN (k = 13) | 0.6139 | 0.1824 |
| Catboost | 0.5954 | 0.2163 |

Although the training results have differed between models, the (linear regression was the best model), it is also important to check the results for the test phase to define the best model to be used.

The tests were performed on the dataset with 33 rows not yet seen by the trained models. Linear Regression and Catboost performed very similarly. Both graphics were also assessed, and it was possible to notice a better fit to the trends, which was explained on a slightly better standard deviation. The test results are shown in Table 2.

**Table 2: Test models results**

| Test phase – Best Models | RMSE | Standard Deviation |
|---|---|---|
| Linear Regression | 0.5435 | 0.3665 |
| KNN (k = 13) | 0.5946 | 0.3725 |
| Catboost | 0.5506 | 0.3327 |

Due to all exposed, considering the smaller standard deviation, the Catboost model with the below parameters was chosen as the best model for the proposed problem solution:

- Catboost Regressor
- loss_function = 'RMSE'
- depth = 6
- l2_leaf_reg = 7
- learning_rate = 0.1

## 4 Conclusions

The main goal of this work was to correctly predict the rating on IMDb based on the main topic of the episode, its director, number of viewers, and number of reviewers. To do so, it's necessary to get variables that strongly correlate with the dependent one (the rating).

Unfortunately, all the independent variables had little direct influence on the rating in our case. Nevertheless, with the features of each episode, it was possible to predict IMDb ratings with an acceptable root mean squared error of 0.55. The model that performed the best to predict the rating was Catboost due to a similar mean squared error of the KNN model but a better standard deviation during the test phase.

The topics generated by the LDA model are difficult to evaluate because, ultimately, they are defined by humans (in this case, the authors), and it is open for interpretation. In this dataset, it is clear that the director's name of the episode has little impact on the rating. The topics about generalist episodes and romantics ones, contrary to what was expected, had a low positive correlation with the variable to be predicted and the episodes that involved family plots had an even lower correlation.

As there was a low correlation between the predictor and its independent variables, some enhancements could be implemented as part of future work, for instance, text mining. There are enhancement possibilities on clusters' number, word filtering (e.g., words with fewer than three characters, etc.), and some counters of words that are considered important or influence the Tv series quality.

Another possibility is to implement complementary technics to pull out more information from scripts such as TF-IDF (Term Frequency — Inverse Document Frequency). TF-IDF is a numerical statistic method that determines the weight, which evaluates the importance of terms (or words) in document collection [16].

Also, the dataset was considered small, which influenced the model selection, leading to models with less dependence on the dataset size. Unfortunately, there was no possibility of increasing the dataset. Nevertheless, it might be possible to apply a similar approach to predict other TV series with more episodes than the current study.

Furthermore, a more generalist model would be interesting to predict what most impacts the popularity of action series. By generating themes for the episodes with the best ratings, it is possible to conduct a feature importance analysis to observe what the audience likes most.

All the resources used throughout this article, like the Python scripts and the dataset are available in our GitHub repository: https://github.com/annalug/Webscraping_arrow_scripts_LDA_topic_modelling.